\documentclass{ieeetj}
\pdfoutput=1

\usepackage{cite}
\usepackage{amsmath,amssymb,amsfonts}
\usepackage{graphicx,color}
\usepackage{textcomp}
\usepackage{xcolor}
\usepackage{hyperref}
\hypersetup{hidelinks=true}

\usepackage{bm}
\usepackage{theorem}
\usepackage{times}
\usepackage{subcaption}
\usepackage{url}
\usepackage[capitalise]{cleveref}
\usepackage{tikz, pgfplots}
\usepackage[ruled,linesnumbered]{algorithm2e}
\usepackage{moresize} 

\def\BibTeX{{\rm B\kern-.05em{\sc i\kern-.025em b}\kern-.08em
    T\kern-.1667em\lower.7ex\hbox{E}\kern-.125emX}}
\AtBeginDocument{\definecolor{tmlcncolor}{cmyk}{0.93,0.59,0.15,0.02}\definecolor{NavyBlue}{RGB}{0,86,125}}
\def\OJlogo{\vspace{-4pt}$<$Society logo(s) and publication title will appear here.$>$}
\def\seclogo{\vspace{10pt}$<$Society logo(s) and publication title will appear here.$>$}
\def\authorrefmark#1{\ensuremath{^{\textbf{#1}}}}

\input{mysymbol.sty}

\newcommand{\padd}[2]{\underline{#1}^{(#2)}}
\def\bchkS{{\ensuremath{\mathbf{\check S}}}}

\def \ubS{\underline{\bbS}}
\def \ubI{\underline{\bbI}}
\def \uccalS{\underline{\ccalS}_{N_t}}
\def \uccalZ{\underline{\ccalZ}_{N_t}}
\def \uf{\underline{f}}
\def\bchkS{{\ensuremath{\mathbf{\check S}}}}

\usetikzlibrary{shapes,arrows}
\pgfplotsset{compat=1.17}
\pgfplotstableset{col sep=semicolon}
\usepgfplotslibrary{groupplots}
\usepgfplotslibrary{fillbetween}

\tikzset{every mark/.append style={scale=1.6, solid}, font=\small}
\pgfplotsset{
    width=1\textwidth,
    height=5.5cm,
    legend style={
        font=\small ,  
        inner xsep=1pt,
        inner ysep=1pt,
        nodes={inner sep=1pt}},
    legend cell align=left,
    every axis/.append style={line width=.5pt},
 	every axis plot/.append style={line width=1.5pt},
 	every axis y label/.append style={yshift=-4pt}
}

\begin{document}
\receiveddate{XX Month, XXXX}
\reviseddate{XX Month, XXXX}
\accepteddate{XX Month, XXXX}
\publisheddate{XX Month, XXXX}
\currentdate{XX Month, XXXX}
\doiinfo{XXXX.2022.1234567}

\markboth{}{Author {et al.}}

\title{Online Learning Of Expanding Graphs}

\author{Samuel Rey\authorrefmark{1} (Member, IEEE), Bishwadeep Das\authorrefmark{2} (Student Member, IEEE), \\ and Elvin Isufi \authorrefmark{2}, (Senior Member, IEEE)}
\affil{Department of Signal Theory and Communications, King Juan Carlos University, Madrid, Spain}
\affil{Intelligent Systems Department, Delft University of Technology, Delft, The Netherlands}
\corresp{Corresponding author: Samuel Rey (email: samuel.rey.escudero@urjc.es).}
\authornote{This work was partially supported by the the Spanish AEI Grants PID2022-136887NB-I00, TED2021-130347B-I00, PID2023-149457OB-I00, the EU H2020 Grant Tailor (No 952215, agreement 99), the Community of Madrid (Madrid ELLIS Unit). TU Delft AI Labs programme, the NWO OTP GraSPA proposal \#19497, and the NWO VENI proposal 222.032.}

\begin{abstract}
This paper addresses the problem of online network topology inference for expanding graphs from a stream of spatiotemporal signals. Online algorithms for dynamic graph learning are crucial in delay-sensitive applications or when changes in topology occur rapidly. While existing works focus on inferring the connectivity within a fixed set of nodes, in practice, the graph can grow as new nodes join the network. This poses additional challenges like modeling temporal dynamics involving signals and graphs of different sizes. This growth also increases the computational complexity of the learning process, which may become prohibitive. To the best of our knowledge, this is the first work to tackle this setting. We propose a general online algorithm based on projected proximal gradient descent that accounts for the increasing graph size at each iteration. Recursively updating the sample covariance matrix is a key aspect of our approach. We introduce a strategy that enables different types of updates for nodes that just joined the network and for previously existing nodes. To provide further insights into the proposed method, we specialize it in Gaussian Markov random field settings, where we analyze the computational complexity and characterize the dynamic cumulative regret. Finally, we demonstrate the effectiveness of the proposed approach using both controlled experiments and real-world datasets from epidemic and financial networks.
\end{abstract}



\maketitle

\section{Introduction}
\IEEEPARstart{N}{etwork} topology inference is a relevant problem aiming at inferring the underlying irregular structure of the observed data~\cite{mateos2019connecting}.
A gamut of methods have been developed to identify \emph{static} graphs with fixed sets of nodes and edges by exploiting different priors that link the graph structure to the properties of the data~\cite{kalofolias2016learn,segarra2017network,dong2019learning,coutino2020state,buciulea2022learning,rey2022joint,zhang2024graph,navarro2024joint,buciulea2024polynomial}.
Nevertheless, graphs change with time~\cite{natali2022learning,das2024tensor}.
Indeed, dynamic topology identification accounts for changes in the support of the graph, which are assumed to be smooth over time.
One research direction in this area harnesses prior knowledge of the graph temporal variations and assumes all data is simultaneously available in batches~\cite{hallac2017network,cirillo2023learning,yamada2020time,kalofolias2017learning}.
An important limitation of these methods is that, in time-varying applications, data often arrives sequentially over time, posing difficulties to batch-based methods.
Moreover, estimating the graph with high accuracy at each time instant might be computationally prohibitive, especially in delay-sensitive applications that require rapid updates of the graph.

Online topology inference methods emerge to overcome the previous limitations.
These efficient algorithms, tailored to processing streaming data, update the estimated graph topology with each incoming observation.
Mimicking the offline case, online topology inference methods are based on
smoothness~\cite{saboksayr2023dual,zhang2022online,saboksayr2021online},~graph stationarity \cite{shafipour2020online}~vector autoregressive models~\cite{money2022online,money2023sparse,zaman2023online}~structural equation models~\cite{natali2022learning}~heat diffusion,~\cite{vlaski2018online}~or non-linear models~\cite{moscu2020online}.
On top of this, new alternatives are also based on prediction correction methods~\cite{natali2022learning}.

Although relevant, the previous methods assume that the set of nodes remains fixed over time.
However, in dynamic scenarios, the size of the graph often increases as new nodes join the network.
Consider for instance a pandemic like COVID-19, where only a few cities initially tracked infection counts but, as the virus spread, more cities and counties began reporting cases, adding new nodes to the epidemic spread network.
Discovering the connections between newly added nodes and the existing network can be crucial for predicting future case counts and modeling the propagation of the disease.
In a different context, expanding graphs are also prevalent in recommended systems, where new users are constantly joining the network~\cite{isufi2021accuracy,das2022task}, or in financial data~\cite{de2020learning}, where stocks related to new companies appear in the market.

Recent studies are starting to tackle problems involving expanding graphs.
A line of works explores processing signals defined on these growing graphs by leveraging prior information about the connectivity of incoming nodes~\cite{venkitaraman_recursive_2020,jian2018toward,dornaika2017efficient,cervino2023learning,das2022online}.
In more general scenarios where the connectivity of new nodes is unknown, some methods approximated it via stochastic priors or inferred it from a task-dependent perspective~\cite{das2022task,barabasi_emergence_1999,erdos_evolution_1961}. 
However, these works focus on processing data by assuming the graph is known.

Departing from previous approaches, we learn the topology of expanding graphs from spatio-temporal streaming signals.
The increasing dimension of the graph involves processing a stream of signals with time-varying dimensions and comparing graphs of different sizes to model the temporal dynamics.
On top of this, as the graph grows, the computational complexity can become prohibitive, underscoring the need for efficient algorithms.
To overcome these challenges, we propose an online algorithm with low computational complexity based on projected proximal gradient (PPG) descent, capable of tracking changes in the topology by performing a few iterations with each new observation.
A key feature of our method is leveraging the block structure of the graph, enabling us to compare the changes in the topology over time.

After reviewing basic ideas of network topology inference in \cref{s:preliminaries}, our main contributions and the structure of the paper are summarized next:
\begin{itemize}
    \item We study the structure resulting from the expanding graph setting and formulate an offline optimization problem to learn the dynamic topology in \cref{s:problem_formulation}.

    \item We proposed an online topology inference algorithm tailored to learn expanding graphs in \cref{s:online_learning}.
    The flexible formulation can accommodate different relations between the graph and the signals.

    \item We specialize our online algorithm to deal with Gaussian Markov random field (GMRF) observations, analyzing its computational complexity and dynamic cumulative regret in \cref{s:online_ggm}.
\end{itemize}
We demonstrate the effectiveness of the proposed method in \cref{s:experiments} over controlled and real-world datasets, followed by concluding remarks in \cref{s:conclusions}.

\section{Fundamentals of network topology inference}\label{s:preliminaries}

The problem of learning a graph involves inferring its topology from a set of nodal observations~\cite{mateos2019connecting}.
Specifically, let $\ccalG = (\ccalV, \ccalE)$ denote a graph with a node set $\ccalV$ consisting of $N$ nodes and an edge set $\ccalE \subseteq \ccalV \times \ccalV$.
Let $\bbS \in \reals^{N \times N}$ be the graph-shift operator (GSO), a matrix representing the graph structure such as the adjacency matrix $\bbA$ or the graph Laplacian $\bbL$~\cite{ortega2018graph,rey2023robust,isufi2024graph}.
Nodal observations are represented by \emph{graph signals}, which are defined on the set $\ccalV$ and expressed as a vector $\bbx \in \mathbb{R}^N$, where $x_i$ denotes the signal value at node $i$.

Given a matrix of $T$ graph signals $\bbX := [\bbx_1, \dots, \bbx_T] \in \mathbb{R}^{N \times T}$, the graph learning problem comprises solving the inverse problem $\bbS = F^{-1}(\bbX)$, where $F(\cdot)$ is a function that relates the GSO to the observed signals.
Consequently, the success of this task hinges on the dependence between $\bbX$ and the unknown $\bbS$, with different methods leveraging various relations.
For instance, graphical models~\cite{ravikumar2011high} assume $\bbS = \bbC^{-1}$, where $\bbC$ denotes the covariance of the observed data; graph stationarity~\cite{segarra2017network} posits that $\bbC$ is a polynomial of $\bbS$; and signal smoothness~\cite{kalofolias2016learn} requires the observed signals to be smooth over the graph.

Denoting the sample covariance matrix as $\hbC = \frac{1}{T}\bbX\bbX^\top$, the graph topology can be recovered by solving
\begin{equation}\label{e:static_nti}
    \bbS^* = \argmin_{\bbS \in \ccalS}   \ccalL(\bbS, \hbC_t) + \lambda r(\bbS),
\end{equation}
where $\ccalL$ is a loss function encoding the relation between $\bbS$ and the observed signals, $r(\bbS)$ is a regularizing promoting desired properties on $\bbS$, and $\ccalS$ defines the set of constraints for a valid GSO, like the non-negative adjacency matrix or graph Laplacian.
Here, we focus on $r(\bbS) = \| \bbS \|_1$ to enforce graph sparsity, but alternative penalties may be considered.

\vspace{2mm}
\noindent
\textbf{Learning graphs from streaming data.}
In many applications, graph signals arrive sequentially over time.
Formally, let $\bbx_t$ denote the signal observed at time $t$, leading to the observation of a sequence $ \{ \bbx_t \}_{t=1}^T$.
To mitigate memory issues associated with large values of $T$, it is common practice to recursively update the covariance matrix $\hbC_t$.
Upon the arrival of a new datum $\bbx_t$, common methods to update $\hbC_{t-1}$ include the rank-one correction 
\begin{equation}\label{e:stationary_update}
    \hbC_t = \frac{t-1}{t} \hbC_{t-1} + \frac{1}{t}\bbx_t\bbx_t^\top,
\end{equation}
tailored to stationary settings~\cite{shafipour2020online}, or the exponential decay
\begin{equation}\label{e:dynamic_update}
    \hbC_t = \gamma \hbC_{t-1} + (1-\gamma) \bbx_t\bbx_t^{\top},
\end{equation}
employed in non-stationary settings \cite{natali2022learning}, where $\gamma \in [0, 1)$ is the forgetting factor.
These types of updates are prevalent in online algorithms where \eqref{e:stationary_update} is solved on the fly~\cite{shafipour2020online,natali2022learning}.

\section{Learning expanding graphs}\label{s:problem_formulation}

\begin{figure}[t]
    \centering
    {\includegraphics[trim=10 5 0 0,clip,width=0.55\textwidth]{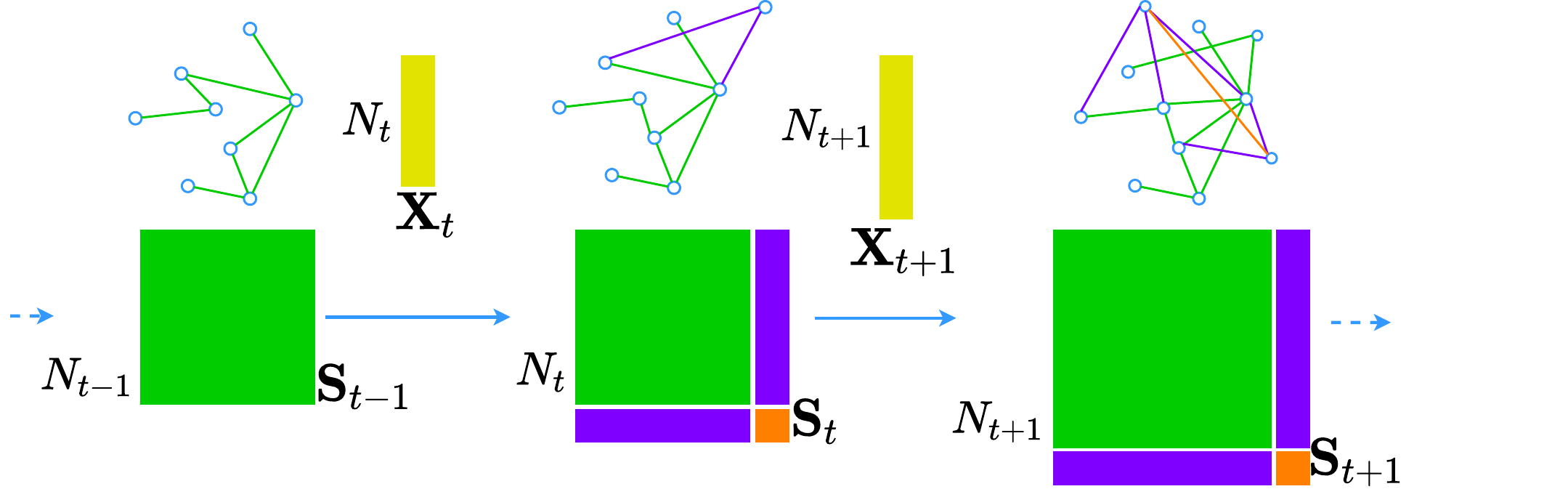}}%
\caption{Evolution of $\ccalG_t$ and $\bbS_t$ in the expanding graph setting. 
Green edges and blocks of the GSO denote the connectivity of previous nodes in the network while purple indicates the connectivity of incoming nodes.
Starting at $t-1$, the initial GSO $\bbS_{t-1}$ is an $N_{t-1}\times N_{t-1}$ matrix.
At time $t$, a new signal $\bbx_t \in \reals^{N_t}$ (yellow) arrives and a new node joins the network, enlarging the size of the graph. Similarly, two nodes arrive on time $t+1$ and so on.
}
\label{f:general_illustration}
\vspace{-5mm}
\end{figure}

This work addresses the problem of learning the topology of a dynamic graph whose set of nodes increases with time.
The evolving topology can be represented by the sequence $\{ \ccalG_t \}_{t=1}^T$.
In the context of expanding graphs, new nodes join the network over time.
Denoting the incoming nodes at time $t$ as $\ccalI_t$, this leads to the node sets $\ccalV_t = \ccalV_{t-1} \cup \ccalI_t$, with $N_t \geq N_{t-1}$.
The growing dimensionality of the node set poses additional challenges when relating graphs and data across different time instants, a critical element in learning time-varying graphs. 

Regarding the unknown topology of the dynamic graph, $\ccalG_t$ is related to its GSO $\bbS_t$, represented by an $N_t \times N_t$ matrix.
Since the nodes are added sequentially over time, each $\bbS_t$ exhibits the following block structure
\begin{equation}\label{e:expanding_GSO}
    \bbS_t = 
    \begin{bmatrix}
        [\bbS_t]_{\ccalV_{t-1}} & [\bbS_t]_{\ccalV_{t-1},\ccalI_t} \\
        [\bbS_t]_{\ccalI_t,\ccalV_{t-1}} & [\bbS_t]_{\ccalI_t}
    \end{bmatrix}.
\end{equation}
Here, the block $[\bbS_t]_{\ccalV_{t-1}} \in \reals^{N_{t-1} \times N_{t-1}}$ encodes the connectivity among nodes present in the network at $t-1$.
Similarly, the blocks $[\bbS_t]_{\ccalV_{t-1},\ccalI_t}$ and $[\bbS_t]_{\ccalI_t,\ccalV_{t-1}}$ respectively denotes the connectivity between incoming and previously present nodes, and $[\bbS_t]_{\ccalI_t}$ encodes the connectivity among incoming nodes.
This block structure allows us to relate GSOs across different time instants by directly comparing $\bbS_{t-1}$ with the block $[\bbS_t]_{\ccalV_{t-1}}$.
Moreover, it helps in capturing prior information about the connectivity involving new nodes.
For instance, if nodes in $\ccalI_t$ can only have edges with nodes in $\ccalV_{t-1}$ we have $[\bbS_t]_{\ccalI_t} = \bbzero$.
Similarly, if the connectivity of nodes in $\ccalV_{t-1}$ remains unchanged we have $[\bbS_t]_{\ccalV_{t-1}} = \bbS_{t-1}$.
The sequence of expanding graphs and the resulting block structure of $\bbS_t$ is depicted in \cref{f:general_illustration}.

With these in place, we formalize the problem at hand.
\begin{problem}\label{p:problem}
    \vspace{-3mm}
    Given a sequence of graph signals $\{ \bbx_t \}_{t=1}^T$, with $\bbx_t \in \reals^{N_t}$ available at time $t$, our aim is to learn the dynamic topology encoded in the sequence of GSOs $\{ \bbS_t \}_{t=1}^T$, under the assumptions: \\
    \noindent\textbf{(AS1)}  
    The set of nodes grows with time, i.e., $\ccalV_{t-1} \subseteq \ccalV_t$. \\
    \noindent\textbf{(AS2)}
    $[\bbS_t]_{\ccalV_{t-1}}$ and $\bbS_{t-1}$ are close in some distance $d(\cdot, \cdot)$.
    
\end{problem}

Here, \textbf{(AS1)} captures the key characteristic of the expanding graph scenario by accounting for new nodes joining the network over time.
\textbf{(AS2)} ensures that the connectivity between nodes already in the network (modeled at time $t$ by the block $[\bbS_t]_{\ccalV_{t-1}}$) changes smoothly.
This assumption makes the topology inference problem tractable and ensures that the main discrepancies between $\bbS_{t-1}$ and $\bbS_t$ are primarily due to the addition of new nodes~\cite{kalofolias2017learning,hallac2017network}.

Given \cref{p:problem}, the time-varying network can be estimated at each time $t$ in an offline manner by solving
\begin{equation}\label{e:offline_nti}
    \bbS^*_t =  \;\argmin_{\bbS \in \ccalS_{N_t}} \ccalL(\bbS, \hbC_t) + \alpha d([\bbS]_{\ccalV_{t-1}}, \bbS^*_{t-1}) + \lambda \| \bbS \|_1.
\end{equation}

Similar to the static problem in \eqref{e:static_nti}, the term $\ccalL(\bbS, \hbC_t)$ represents a loss function of interest relating the graph topology and the signal (statistical) properties.
Note that the dimension of $\hbC_t$ increases with time, so the ensuing section introduces an update strategy tailored to expanding graphs.
Next, recalling that $[\bbS]_{\ccalV_{t-1}}$ encodes the topology related to nodes already present at time $t-1$, $d([\bbS]_{\ccalV_{t-1}}, \bbS_{t-1}^*)$ captures any prior knowledge about the evolution of the graph with time.
For instance, if the weights of the edges change across time but the support remains fixed, a Frobenius penalty might be adequate.
In contrast, an $\ell_1$ norm penalty might be preferred if changes only involve rewiring a few edges.
Finally, the set of constraints $\ccalS_{N_t}$ reflects that the dimension of $\bbS^*_t$ varies with time. 

Solving the offline problem \eqref{e:offline_nti} can provide a high-quality estimate $\bbS_t^*$.
Nonetheless, many delay-sensitive applications where new data arrives rapidly may not allow obtaining such an estimate since it may incur excessive computational costs.
In settings where changes in the topology of $\bbS_t$ take place with high frequency, it might not even be prudent to solve \eqref{e:offline_nti} with high precision.
Indeed, this poses a trade-off where sub-optimal and fast solutions may be preferred over optimal but time-consuming alternatives. 
Motivated by these limitations, we introduce next an online low-complexity algorithm to solve \cref{p:problem}.

\section{Online learning of expanding graphs}\label{s:online_learning}
Online graph learning methods are well-suited for learning dynamic graphs from streaming data \cite{saboksayr2021online,money2022online}.
Here, we first study how to update the sample covariance matrix and then introduce an algorithm to solve \eqref{e:offline_nti} online.

\subsection{Expanding covariance update}\label{s:cov_update}
When dealing with expanding graphs the dimensions of $\bbx_t$ and $\hbC_{t-1}$ may not match due to the arrival of new nodes, hence rendering classical updates invalid.
We leverage zero-padding to keep track of $\hbC_t$ in an expanding fashion.

For any matrix $\bbZ \in \reals^{i \times i}$, let $\padd{\bbZ}{j} \in \reals^{j \times j}$ be given by
\begin{equation}\label{e:zero_padding}
    \padd{\bbZ}{j} = \begin{bmatrix}
        \bbZ & \bbzero \\
        \bbzero & \bbzero
     \end{bmatrix},
\end{equation}
with $j \geq i$ and each $\bbzero$ being the all zero matrix of appropriate size.
Simply put, $\padd{\bbZ}{j}$ is a matrix of size $j \times j$ constructed by zero-padding $\bbZ$.
Then, consider the update 
\begin{equation}\label{e:c_update}
    \hbC_t = \bbM_1 \circ \padd{\hbC}{N_t}_{t-1} + \bbM_2 \circ \bbx_t \bbx_t^\top,
\end{equation}
where $\circ$ denotes the entry-wise product, and the masks $\bbM_1$ and $\bbM_2$ are given by
\begin{align}\label{e:mask_update}
    &\bbM_1 = \begin{bmatrix}
        \gamma \bbone & \frac{\tilde{t}-1}{\tilde{t}} \bbone \\
        \frac{\tilde{t}-1}{\tilde{t}} \bbone & \frac{\tilde{t}-1}{\tilde{t}} \bbone
    \end{bmatrix},
    &\bbM_2 = \begin{bmatrix}
         (1-\gamma) \bbone & \frac{1}{\tilde{t}} \bbone \\
         \frac{1}{\tilde{t}} \bbone & \frac{1}{\tilde{t}} \bbone
     \end{bmatrix},
\end{align}
with $\tilde{t} = t - \tau$, $\tau$ being the last time instance that new nodes arrived, and the top left blocks of $\bbM_1$ and $\bbM_2$ being matrices of size $N_\tau \times N_\tau$.
Masks $\bbM_1$ and $\bbM_2$ help in using different update strategies for the blocks in $\hbC_t$ related to new nodes and the blocks related to previously existing nodes.
The motivation for this update is twofold.
First, it allows us to keep track of the changes in $\hbC_{t-1}$ due to the addition of new nodes via a dynamic update for the top left block of $\hbC_t$.
At the same time, the temporal stationary update of the blocks corresponding to new nodes accounts for the lack of prior observations, thereby helping to gather sufficient data from these nodes faster.

\subsection{Online algorithm}
We now propose an online algorithm based on PPG descent, a well-established method for minimizing non-smooth functions such as the $\ell_1$ norm present in our objective function~\cite{beck2009fast,parikh2014proximal}.
For simplicity, we assume that the distance metric $d(\cdot, \cdot)$ is differentiable and group the smooth terms of the objective function in \eqref{e:offline_nti} as 
\begin{equation}\label{e:smooth_terms}
    f_t(\bbS) =  \ccalL(\bbS, \hbC_t) + \alpha d([\bbS]_{\ccalV_{t-1}}, \hbS_{t-1}).
\end{equation}
Nonetheless, non-differentiable distance functions can also be accommodated by modifying the proximal operator.

As summarized in Algorithm~\ref{a:nti_expanding}, our iterative algorithm performs different steps at each time $t$.
These steps involve: i)~updating $\hbC_t$; ii)~performing a PPG step; and iii)~combining the estimate from $t-1$ and the result of the second step.
This process is detailed next.

\vspace{1mm}
\noindent\textbf{Step 1.}
Upon receiving a new observation $\bbx_t$, we update the sample covariance $\hbC_t$ according to \eqref{e:c_update}.

\vspace{1mm}
\noindent\textbf{Step 2.}
We take a PPG step with respect to the function \eqref{e:smooth_terms}, given by
\begin{equation}
    \bchkS_t = \Pi_{\ccalS_{N_t}} \left( T_{\eta\lambda} \left( \padd{\hbS}{N_t}_{t-1} - \eta \nabla f(\padd{\hbS}{N_t}_{t-1}) \right) \right).
\end{equation}
The proximal operator of the $\ell_1$ norm is the soft-thresholding operator, which can be computed entry-wise as $T_{\lambda} (Z_{ij}) = \max \{ |Z_{ij}| - \lambda, 0 \} \sign( Z_{ij} )$ for some matrix $\bbZ$. The scalar $\eta > 0$ represents the step size, $\Pi_{\ccalS_{N_t}}$ denotes the projection onto the feasible set $\ccalS_{N_t}$, $\hbS_{t-1}$ is the estimated GSO at the previous iteration, and $\padd{\hbS}{N_t}_{t-1}$ denotes its zero-padded version as defined in \eqref{e:zero_padding}. 
This zero-padding strategy ensures the dimension of $\hbS_t$ grows dynamically as new nodes arrive.

\vspace{1mm}
\noindent\textbf{Step 3.}
The estimate of the GSO at time $t$ is given by the convex combination
\begin{equation}
    \hbS_t = h \bchkS_t + (1-h) \padd{\hbS}{N_t}_{t-1},
\end{equation}
controlled by the scalar $h \in (0, 1]$.
Values of $h$ close to 0 promote solutions that vary slowly with respect to the previous estimate, while $h=1$ focuses exclusively on the estimate resulting from the proximal gradient step.
Analogous to the covariance update in \eqref{e:c_update}, different values of $h$ can be applied to blocks corresponding to nodes in $\ccalV_{t-1}$ and in $\ccalI_t$.

\begin{algorithm}[tb]
\footnotesize
\SetKwInput{Input}{Input}
\SetKwInOut{Output}{Output}
\Input{Sequence $\{ \bbx_t \}_{t=1}^T$, function $f_t$, stepsize $\eta$, and $h$.}
\SetAlgoLined
    Initialize $\hbS_0 \in \ccalS_0$. \\
   \For{$t=1$ \KwTo $T$}{
        Update the masks $\bbM_1$ and $\bbM_2$ as in \eqref{e:mask_update}. \\
        Covariance update: {\small $\hbC_t = \bbM_1 \circ \padd{\hbC}{N_t}_{t-1} + \bbM_2 \circ \bbx_t\bbx_t^\top$}. \\
        PPG step:
        {\small $ \bchkS_t = \Pi_{\ccalS_{N_t}} \left(  T_{\eta\lambda} \left( \padd{\hbS}{N_t}_{t-1} - \eta \nabla f_t(\padd{\hbS}{N_t}_{t-1}) \right) \right)$}. \\
        Update estimate: {\small $\hbS_t = h \bchkS_t + (1-h) \padd{\bbS}{N_t}_{t-1}$}.
        }
        \Output{Estimated sequence $\{ \hbS_t \}_{t=1}^T$.}
        \caption{\small Online graph learning over expanding graphs.}\label{a:nti_expanding}
\end{algorithm}

Algorithm~\ref{a:nti_expanding} considers a single proximal gradient iteration per time instant, a design tailored to highly delay-sensitive applications.
Additional iterations can be performed if a more accurate estimate is preferred over obtaining a fast solution.
This application-dependent trade-off is determined by the frequency of which new nodes and data arrive and is studied in detail in \cref{s:experiments}.
Finally, note that the proposed algorithm considers a general loss function $\ccalL$ linking the data to the underlying topology, with the only requirement being its differentiability.
Consequently, we can use Algorithm~\ref{a:nti_expanding} to learn the topology of expanding graphs for different graph-data priors, such as GMRF~\cite{rey2023enhanced}, graph stationarity~\cite{segarra2017network}, or structural equation models~\cite{baingana2014proximal}, to name a few. 
The following section studies the particular case where the observed signals follow a GMRF.

\section{Online expanding Gaussian graphical model}\label{s:online_ggm}
In Gaussian graphical models, each observation $\bbx_t \in \reals^{N_t}$ is sampled from a zero-mean GMRF, implying $\bbx_t$ adheres to the zero-mean multivariate Gaussian distribution with covariance $\bbC_t = \bbS_t^{-1}$, whose dimension increases with $t$ due to the growing graph size.
Here, the graph topology, or equivalently, the inverse of the covariance matrix, is learned by maximizing the (regularized) log-likelihood of the observed data~\cite{rey2023enhanced}.

The maximum likelihood estimator of the GSO at time $t$ can be computed offline by solving 
\begin{alignat}{2}\label{e:offline_ggl}
    \!\!&\!  \bbS^*_t =    
    && \;\argmin_{\bbS}  \;\; \tr(\bbS\hbC_t) - \log\det(\bbS + \epsilon\bbI) + \lambda \| \bbS \|_1  \\ 
    \!\!&\!  &&  \qquad\qquad + \alpha d([\bbS]_{\ccalV_{t-1}}, \bbS^*_{t-1})  \nonumber \\
    \!\!&\!  && \mathrm{s. \;t. } \;\bbS \in \ccalS_{N_t} := \{ \bbS \in \reals^{N_t \times N_t} | \bbS \succeq 0, ~ \| \bbS \|_2^2 \leq \sigma \}.  \;\; \nonumber
\end{alignat}
Similar to existing works on graph Laplacian Gaussian graphical models~\cite{ying2020nonconvex}, the term $\epsilon\bbI$ for $\epsilon > 0$ in the determinant, allows us to obtain positive semi-definite GSOs, thus enhancing the practicality of our method.
The ability to estimate rank-deficient matrices allows for disconnected graph solutions.
Moving on to the set $\ccalS_{N_t}$, the constraint $\bbS \succeq 0$ ensures $\bbS$ is a positive semidefinite matrix and $ || \bbS ||_2^2 \leq \sigma $ imposes bounded eigenvalues, as is typically the case in real-world networks.
This set also reflects the growing size of the graph.
The distance term $d(\cdot, \cdot)$ captures \textbf{(AS2)} by modeling how the connectivity of nodes previously existing in the network evolves. 
When no prior information about the relation between $[\bbS]_{\ccalV_{t-1}}$ and $\bbS_{t-1}$ is available, it suffices to set $\alpha = 0$.

We solve \eqref{e:offline_ggl} online using Algorithm~\ref{a:nti_expanding}.
Let $f_t(\bbS)$ encompass all the terms in the objective except for the $\ell_1$ norm, resulting in the gradient
\begin{equation}\label{e:gradient_ggl}
    \nabla f_t(\bbS) = \hbC_t - \left( \bbS + \epsilon\bbI \right)^{-1} + \alpha \nabla d([\bbS]_{\ccalV_{t-1}}, \hbS_{t-1}).
\end{equation}
Since $d(\cdot, \cdot)$ only depends on the entries of $\bbS$ related to nodes in $\ccalV_{t-1}$, note that $\nabla d(\cdot, \cdot)$ will be an $N_t \times N_t$ matrix with zeros in the entries not related nodes in $\ccalV_{t-1}$. 
It can be shown that $f_t(\bbS)$ is strongly convex with constant $\sigma^{-1}$ and its gradient is Lipschitz continuous with constant $\epsilon^{-2}$ (given that $\nabla d$ is also Lipschitz)~\cite{navarro2024fair}.
Therefore, we select the step size such that $\eta \leq \epsilon^2$.
The projection onto $\ccalS_{N_t}$ can be computed as
\begin{equation}\label{e:projection_ggl}
    \Pi_{\ccalS_{N_t}}(\bbS) = \bbV \min \left\{ \max \left\{ \bbLambda, 0 \right\}, \sigma^{1/2} \right\} \bbV^\top,
\end{equation}
where $\bbV$ and $\bbLambda$ denote the eigenvectors and eigenvalues of $\bbS$, i.e., $\bbS = \bbV \bbLambda \bbV^\top$.
With some abuse of notation, we let $\min \{ \max\{ \bbLambda, 0 \}, \sigma^{1/2} \}$ represent the element-wise minimum and maximum operation on the entries of $\bbLambda$.
This operation projects the eigenvalues contained in the diagonal of $\bbLambda$ into the interval $[0, \sigma^{1/2}]$.

\vspace{1mm}
\noindent\textbf{Computational complexity.}
From a computational standpoint, the complexity of each iteration of Algorithm~\ref{a:nti_expanding} is dictated by the matrix inverse in \eqref{e:gradient_ggl} and the eigendecomposition in \eqref{e:projection_ggl}.
While over-the-shelf implementations of these operations incur a computational complexity of $\ccalO(N_t^3)$, efficient implementations based on fast matrix multiplications can reduce the complexity to $\ccalO(N_t^{2.4})$~\cite{williams2024new}.
We stress that a complexity of $\ccalO(N_t^{2.4})$ is considerably efficient since learning the graph topology involves $\ccalO(N_t^2)$ variables.

\subsection{Dynamic cumulative regret analysis}
An important question in online graph learning is whether Algorithm~\ref{a:nti_expanding} closely tracks the sequence of minimizes $\{ \bbS^*_t \}_{t=1}^T$ for large values of $T$.
To answer this question, we analyze the dynamic cumulative regret, defined as $ \sum_{t=1}^T \| \hbS_t - \bbS_t^* \|_F$~\cite{mokhtari2016online}.
The expanding graph setting makes our dynamic regret analysis substantially different from classical results in conventional online learning.
The following theorem establishes a bound for the dynamic cumulative regret of the sequence provided by Algorithm~\ref{a:nti_expanding} for the Gaussian graphical models described in \cref{s:online_ggm}.
\begin{theorem}\label{t:convergence}
    Let $\{ \hbS_t\}_{t=1}^T$ be the sequence of estimates procured by Algorithm~\ref{a:nti_expanding} with $\nabla f_t(\bbS)$ and $\Pi_{\ccalS_{N_t}}(\bbS)$ as given in \eqref{e:gradient_ggl} and \eqref{e:projection_ggl}, respectively.
     Let $\{ \bbS_t^* \}_{t=1}^T$ be the sequence of minimizers of \eqref{e:offline_ggl}.
    Setting the step size as $\eta \leq \epsilon^2$, the dynamic cumulative regret is upper bounded by
    \begin{equation}\label{e:theorem1}
        \sum_{t=1}^T \| \hbS_t - \bbS_t^* \|_F \leq K_1 + K_2\sum_{t=2}^T \| \bbS_t^* - \padd{\bbS^*}{N_t}_{t-1}\|_F,
    \end{equation}
    with $\padd{\bbS^*}{N_t}_{t-1}$ being the $N_t \times N_t$ zero-padded version of $\bbS^*_{t-1}$ as defined in \eqref{e:zero_padding}, and with constants
    \begin{align}\label{e:thm_constants}
        &K_1 = \frac{\| \hbS_1 - \bbS_1^* \|_F - \rho \| \hbS_T - \bbS_T^* \|_F}{1-\rho}, &K_2 = \frac{1}{1-\rho},
    \end{align}
    where $0 \leq \rho := (1-h\eta/\sigma)^{1/2} < 1$, $\sigma^{-1}$ is the strong convexity constant of $f_t$, and $h \in (0, 1]$.
\end{theorem}

The theorem is proven in Appendix A.
\cref{t:convergence} asserts that the ability of Algorithm~\ref{a:nti_expanding} to track the dynamic and expanding topology depends on the evolution of the optimal solution to \eqref{e:offline_ggl}, given by $\sum_{t=2}^T \| \bbS_t^* - \padd{\bbS^*}{N_t}_{t-1}\|_F$. This is also known as the path length, which, in turn, reflects the variability of the underlying graph over time.
As $T \to \infty$, \cref{t:convergence} guarantees that if the sum of the variations in $\bbS_t^*$ is sub-linear in time, then $\frac{1}{T} \sum_{t=1}^T || \hbS_t - \bbS^*_t ||_F \to 0$.
To provide additional intuition, consider a scenario where changes in the topology are solely due to the arrival of new nodes with binary edges, while the connectivity of nodes in $\ccalV_{t-1}$ remains unchanged.
Moreover, let $|\ccalI_t|$ and $d_{\textnormal{max}}$ respectively denote the number of incoming nodes and the maximum number of edges per new node.
Then, we have
\begin{equation}
    \| \bbS_t^* - \padd{\bbS^{\star}}{N_t}_{t-1}\|_F \leq \sqrt{2 d_{\textnormal{max}} |\ccalI_t|},
\end{equation}
meaning that the cumulative regret is determined by the total number new edges added to the network.

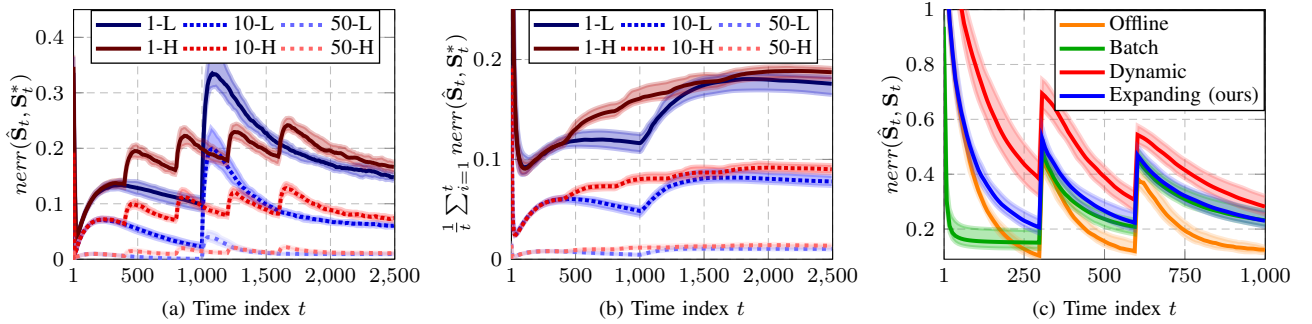
\begin{figure*}[!t]
	\centering
	\begin{subfigure}{0.32\textwidth}
		\centering
		 \begin{tikzpicture}[baseline,scale=.9]

\pgfplotstableread{data/iters/iters_low-err_off_median.csv}\errtablelow
\pgfplotstableread{data/iters/iters_low-err_off_prct75.csv}\prcttoplow
\pgfplotstableread{data/iters/iters_low-err_off_prct25.csv}\prctbotlow
\pgfplotstableread{data/iters/iters_high-err_off_median.csv}\errtablehigh
\pgfplotstableread{data/iters/iters_high-err_off_prct75.csv}\prcttophigh
\pgfplotstableread{data/iters/iters_high-err_off_prct25.csv}\prctbothigh

\pgfmathsetmacro{\opacity}{0.3}
\pgfmathsetmacro{\contourop}{0.25}

\begin{axis}[
    xlabel={(a) Time index $t$},
    xmin=1,
    xmax=2500,
    xtick = {1, 500, 1000, 1500, 2000, 2500},
    ylabel={$nerr(\hbS_t, \bbS^*_t)$},
    ymin = 0,
    ymax = .45,
    ytick = {0, .1, ..., .51},
    grid style=densely dashed,
    grid=both,
    legend style={
        at={(.56, 1.11)},,
        anchor=north},
    legend columns=3,
    width=180,
    height=150,
    ]

    \addplot [blue!80!black, name path = low1-bot, opacity=\contourop, forget plot] table [x=xaxis, y=1-Low] \prctbotlow;
    \addplot [blue!70!black, name path = low1-top, opacity=\contourop, forget plot] table [x=xaxis, y=1-Low] \prcttoplow;
    \addplot[blue!70!black, fill opacity=\opacity, forget plot] fill between[of=low1-bot and low1-top];
    \addplot[blue!40!black, solid] table [x=xaxis, y=1-Low] {\errtablelow};

    \addplot [blue!80!white, name path = low10-bot, opacity=\contourop, forget plot] table [x=xaxis, y=10-Low] \prctbotlow;
    \addplot [blue!90!white, name path = low10-top, opacity=\contourop, forget plot] table [x=xaxis, y=10-Low] \prcttoplow;
    \addplot[blue!90!white, fill opacity=\opacity, forget plot] fill between[of=low10-bot and low10-top];
    \addplot[blue!90!black, densely dotted] table [x=xaxis, y=10-Low] {\errtablelow};
    
    \addplot [blue!40!white, name path = low50-bot, opacity=\contourop, forget plot] table [x=xaxis, y=50-Low] \prctbotlow;
    \addplot [blue!30!white, name path = low50-top, opacity=\contourop, forget plot] table [x=xaxis, y=50-Low] \prcttoplow;
    \addplot[blue!30!white, fill opacity=\opacity, forget plot] fill between[of=low50-bot and low50-top];
    \addplot[blue!60!white, dotted] table [x=xaxis, y=50-Low] {\errtablelow};

    \addplot [red!80!black, name path = high1-bot, opacity=\contourop, forget plot] table [x=xaxis, y=1-High] \prctbothigh;
    \addplot [red!70!black, name path = high1-top, opacity=\contourop, forget plot] table [x=xaxis, y=1-High] \prcttophigh;
    \addplot[red!70!black, fill opacity=\opacity, forget plot] fill between[of=high1-bot and high1-top];
    \addplot[red!40!black, solid] table [x=xaxis, y=1-High] {\errtablehigh};

    \addplot [red!80!white, name path = high10-bot, opacity=\contourop, forget plot] table [x=xaxis, y=10-High] \prctbothigh;
    \addplot [red!90!white, name path = high10-top, opacity=\contourop, forget plot] table [x=xaxis, y=10-High] \prcttophigh;
    \addplot[red!90!white, fill opacity=\opacity, forget plot] fill between[of=high10-bot and high10-top];
    \addplot[red!90!black, densely dotted] table [x=xaxis, y=10-High] {\errtablehigh};
    
    \addplot [red!40!white, name path = high50-bot, opacity=\contourop, forget plot] table [x=xaxis, y=50-High] \prctbothigh;
    \addplot [red!30!white, name path = high50-top, opacity=\contourop, forget plot] table [x=xaxis, y=50-High] \prcttophigh;
    \addplot[red!30!white, fill opacity=\opacity, forget plot] fill between[of=high50-bot and high50-top];
    \addplot[red!60!white, dotted] table [x=xaxis, y=50-High] {\errtablehigh};

    \legend{1-L, 10-L, 50-L, 1-H, 10-H, 50-H}
\end{axis}
\end{tikzpicture}
	\end{subfigure}
	\begin{subfigure}{0.32\textwidth}
		\centering
        \begin{tikzpicture}[baseline,scale=.9]

\pgfplotstableread{data/iters/iters_low-regret_median.csv}\errtablelow
\pgfplotstableread{data/iters/iters_low-regret_prct75.csv}\prcttoplow
\pgfplotstableread{data/iters/iters_low-regret_prct25.csv}\prctbotlow
\pgfplotstableread{data/iters/iters_high-regret_median.csv}\errtablehigh
\pgfplotstableread{data/iters/iters_high-regret_prct75.csv}\prcttophigh
\pgfplotstableread{data/iters/iters_high-regret_prct25.csv}\prctbothigh

\pgfmathsetmacro{\opacity}{0.3}
\pgfmathsetmacro{\contourop}{0.25}

\begin{axis}[
    xlabel={(b) Time index $t$},
    ylabel={$\frac{1}{t}\sum_{i=1}^t nerr(\hbS_t, \bbS_t^*)$},
    xmin=1,
    xmax=2500,
    xtick = {1, 500, 1000, 1500, 2000, 2500},
    ymin = 0,
    ymax = .25,
    ytick = {0, .1, ..., .31},
    grid style=densely dashed,
    grid=both,
    legend style={
        at={(.56, 1.11)},
        anchor=north},
    legend columns=3,
    width=180,
    height=150,
    ]

    \addplot [blue!80!black, name path = low1-bot, opacity=\contourop, forget plot] table [x=xaxis, y=1-Low] \prctbotlow;
    \addplot [blue!70!black, name path = low1-top, opacity=\contourop, forget plot] table [x=xaxis, y=1-Low] \prcttoplow;
    \addplot[blue!70!black, fill opacity=\opacity, forget plot] fill between[of=low1-bot and low1-top];
    \addplot[blue!40!black, solid] table [x=xaxis, y=1-Low] {\errtablelow};

    \addplot [blue!80!white, name path = low10-bot, opacity=\contourop, forget plot] table [x=xaxis, y=10-Low] \prctbotlow;
    \addplot [blue!90!white, name path = low10-top, opacity=\contourop, forget plot] table [x=xaxis, y=10-Low] \prcttoplow;
    \addplot[blue!90!white, fill opacity=\opacity, forget plot] fill between[of=low10-bot and low10-top];
    \addplot[blue!90!black, densely dotted] table [x=xaxis, y=10-Low] {\errtablelow};
    
    \addplot [blue!40!white, name path = low50-bot, opacity=\contourop, forget plot] table [x=xaxis, y=50-Low] \prctbotlow;
    \addplot [blue!30!white, name path = low50-top, opacity=\contourop, forget plot] table [x=xaxis, y=50-Low] \prcttoplow;
    \addplot[blue!30!white, fill opacity=\opacity, forget plot] fill between[of=low50-bot and low50-top];
    \addplot[blue!60!white, dotted] table [x=xaxis, y=50-Low] {\errtablelow};

    \addplot [red!80!black, name path = high1-bot, opacity=\contourop, forget plot] table [x=xaxis, y=1-High] \prctbothigh;
    \addplot [red!70!black, name path = high1-top, opacity=\contourop, forget plot] table [x=xaxis, y=1-High] \prcttophigh;
    \addplot[red!70!black, fill opacity=\opacity, forget plot] fill between[of=high1-bot and high1-top];
    \addplot[red!40!black, solid] table [x=xaxis, y=1-High] {\errtablehigh};

    \addplot [red!80!white, name path = high10-bot, opacity=\contourop, forget plot] table [x=xaxis, y=10-High] \prctbothigh;
    \addplot [red!90!white, name path = high10-top, opacity=\contourop, forget plot] table [x=xaxis, y=10-High] \prcttophigh;
    \addplot[red!90!white, fill opacity=\opacity, forget plot] fill between[of=high10-bot and high10-top];
    \addplot[red!90!black, densely dotted] table [x=xaxis, y=10-High] {\errtablehigh};
    
    \addplot [red!40!white, name path = high50-bot, opacity=\contourop, forget plot] table [x=xaxis, y=50-High] \prctbothigh;
    \addplot [red!30!white, name path = high50-top, opacity=\contourop, forget plot] table [x=xaxis, y=50-High] \prcttophigh;
    \addplot[red!30!white, fill opacity=\opacity, forget plot] fill between[of=high50-bot and high50-top];
    \addplot[red!60!white, dotted] table [x=xaxis, y=50-High] {\errtablehigh};

    \legend{1-L, 10-L, 50-L, 1-H, 10-H, 50-H}
\end{axis}
\end{tikzpicture}
	\end{subfigure}
	\begin{subfigure}{0.32\textwidth}
		\centering
		\begin{tikzpicture}[baseline,scale=.9]

\pgfplotstableread{data/updates/update_50-err_seq_median.csv}\errtable
\pgfplotstableread{data/updates/update_50-err_seq_prct75.csv}\prcttop
\pgfplotstableread{data/updates/update_50-err_seq_prct25.csv}\prctbot

\pgfmathsetmacro{\opacity}{0.3}
\pgfmathsetmacro{\contourop}{0.25}

\begin{axis}[
    xlabel={(c) Time index $t$},
    xmin=1,
    xmax=1000,
    xtick = {1, 250, 500, 750, 1000},
    ylabel={$nerr(\hbS_t, \bbS_t)$},
    ymin = .09,
    ymax = 1,
    grid style=densely dashed,
    grid=both,
    legend style={
        at={(1.11, 1.11)},
        anchor=north east},
    legend columns=1,
    width=180,
    height=150,
    ]

    \addplot [orange!80!white, name path = off-bot, opacity=\contourop, forget plot] table [x=xaxis, y=Off] \prctbot;
    \addplot [orange!70!white, name path = off-top, opacity=\contourop, forget plot] table [x=xaxis, y=Off] \prcttop;
    \addplot[orange!70!white, fill opacity=\opacity, forget plot] fill between[of=off-bot and off-top];
    \addplot[orange, solid] table [x=xaxis, y=Off] {\errtable};

    \addplot [green!70!black, name path = batch-bot, opacity=\contourop, forget plot] table [x=xaxis, y=Batch] \prctbot;
    \addplot [green!70!black, name path = batch-top, opacity=\contourop, forget plot] table [x=xaxis, y=Batch] \prcttop;
    \addplot[green!70!black, fill opacity=\opacity, forget plot] fill between[of=batch-bot and batch-top];
    \addplot[green!65!black, solid] table [x=xaxis, y=Batch] {\errtable};

    \addplot [red!80!white, name path = dyn-bot, opacity=\contourop, forget plot] table [x=xaxis, y=Dyn] \prctbot;
    \addplot [red!70!white, name path = dyn-top, opacity=\contourop, forget plot] table [x=xaxis, y=Dyn] \prcttop;
    \addplot[red!70!white, fill opacity=\opacity, forget plot] fill between[of=dyn-bot and dyn-top];
    \addplot[red, solid] table [x=xaxis, y=Dyn] {\errtable};
    
    \addplot [blue!80!white, name path = exp-bot, opacity=\contourop, forget plot] table [x=xaxis, y=Exp] \prctbot;
    \addplot [blue!70!white, name path = exp-top, opacity=\contourop, forget plot] table [x=xaxis, y=Exp] \prcttop;
    \addplot[blue!70!white, fill opacity=\opacity, forget plot] fill between[of=exp-bot and exp-top];
    \addplot[blue, solid] table [x=xaxis, y=Exp] {\errtable};

    \legend{Offline, Batch, Dynamic, Expanding (ours)}
\end{axis}
\end{tikzpicture}
	\end{subfigure}
    \vspace{-1mm}
	\caption{Evaluating the performance of Algorithm~1 over controlled data. (a) and (b) respectively depict the instantaneous error and the average cumulative regret between the online and the offline solution.
    We consider new groups of nodes arriving with low or high frequency, denoted as ``L'' and ``H'', and perform 1, 10, or 50 iterations of Alfgorithm~1 per time instant.
    (c) illustrates the instantaneous error of various methods relative to the true graph $\bbS_t$.}\label{fig:experiments}
    \vspace{-3mm}
\end{figure*}

Finally, we compare our cumulative regret analysis with related results for online dynamic methods on non-increasing graphs. 
The tracking error bound at time $t$ in \cite{shafipour2020online} depends on the maximum element of the path length up to that time instant.
It is also similar to the prediction correction-based tracking error in \cite[Thm.~4]{natali2022learning} and the dynamic regret of \cite[Thm.~1]{money2023sparse}, both of which also depend on the difference between the optimal solutions over time for fixed sized graphs.
Thus, even with the increasing dimension, we have a tracking error that depends on the path length of the optimal solution which corresponds to those used for identifying the topology in dynamic scenarios.
By extension, the dynamic regret for the online algorithm depends on the path length.

\section{Numerical experiments}\label{s:experiments}
We now assess the performance of the proposed method over controlled and real-world data.
The code with the algorithm and implementation details is available on GitHub\footnote{\url{https://github.com/reysam93/online_ntf_expanding}}.

We measure the performance using the normalized Frobenius error, defined as $nerr(\hbS_t, \bbS_t^*) = \| \hbS_t - \bbS_t^* \|_F^2 / \| \bbS_t^* \|_F^2$, where $\hbS_t$ and $\bbS_t^*$ respectively denote the online and offline estimates at time $t$.
We also consider the average cumulative regret up to time $t$, computed as $\frac{1}{t}\sum_{i=1}^t nerr(\hbS_t, \bbS_t^*)$.

\subsection{Experiments using controlled data}
We begin the numerical evaluation of our algorithm with controlled data, which enables precise performance assessment due to the known ground truth graph, denoted by $\bbS_t$.
Graphs are sampled from an Erd\H{o}s-Rényi model with $100$ nodes and an average node degree of 4, and a single signal $\bbx_t$ sampled from a GMRF arrives at each time $t$.
We report the median along with the 25th and 75th percentiles, computed across 100 independent realizations.

\vspace{2mm}
\noindent
\textbf{Frequency of incoming nodes.}
We investigate how the frequency of incoming nodes affects the quality of our online algorithm.
\cref{fig:experiments}a and \cref{fig:experiments}b illustrate the error and average cumulative regret between the online estimate $\hbS_t$ and the offline solution $\bbS^*_t$, given by \eqref{e:offline_ggl}.
We consider a low-frequency setting (``L'') with one group of 20 nodes arriving simultaneously, and a high-frequency setting (``H'') with 4 groups of 5 nodes arriving at different times. 
Additionally, we perform 1, 10, or 50 iterations of Algorithm~1 per time instant.
The results demonstrate that Algorithm~1 effectively recovers from topology changes due to incoming nodes, even with significant disruptions as in the low-frequency setting.
However, when nodes arrive rapidly, there is not enough time to fully recover, leading to accumulating errors.
Overall, the results align with \cref{t:convergence}, as the impact of the disruption depends on the number of incoming nodes, and the average cumulative regret in \cref{fig:experiments}b converges to a constant.
Furthermore, a larger number of iterations reduce the error at the expense of increasing the running time, which ranges from 2.5 to 91 milliseconds per time instant for 1 to 50 iterations per new sample.

\vspace{2mm}
\noindent
\textbf{Estimation error.}
We analyze the error between the true graph, $\bbS_t$, and the estimate $\hbS_t$, obtained using different methods.
In this scenario, the graph grows due to the arrival of 3 groups of 15 nodes.
\cref{fig:experiments}c illustrates the error $nerr(\hbS_t, \bbS_t)$ when $\hbS_t$ is estimated using: i) ``Offline'': the offline estimated from solving \eqref{e:offline_ggl}; ii) ``Batch'' the iterative batch solution; iii) ``Dynamic'': Algorithm~1 using the standard dynamic covariance update in \eqref{e:dynamic_update} with zero-padding; and iv) ``Expanding'': Algorithm~1 with our covariance update in \eqref{e:c_update} tailored to expanding graphs.
We observe that the batch solution quickly achieves a small error thanks to simultaneously observing all the data but requires more iterations to obtain a performance like the offline solution when the number of samples ($t$) is large enough.
As new nodes arrive, the proposed method attains a performance comparable to ``Batch'', demonstrating the effectiveness of the proposed covariance update.
In contrast, the classical dynamic update requires significantly more iterations to achieve similar results.
Notably, our algorithm requires only a few new samples to recover from the addition of new nodes, displaying a performance close to the offline model.

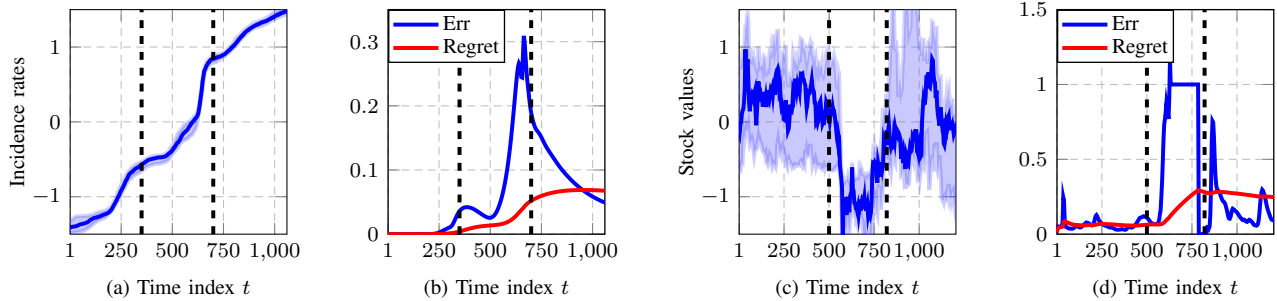
\begin{figure*}[!t]
	\centering
	\begin{subfigure}{0.24\textwidth}
		\centering
		\begin{tikzpicture}[baseline,scale=1]

\pgfplotstableread{data/realdata/covid_incid_med.csv}\errtable
\pgfplotstableread{data/realdata/covid_incid_prct75.csv}\prcttop
\pgfplotstableread{data/realdata/covid_incid_prct25.csv}\prctbot

\pgfmathsetmacro{\opacity}{0.3}
\pgfmathsetmacro{\contourop}{0.25}

\begin{axis}[
    xlabel={(a) Time index $t$},
    xmin=1,
    xmax=1060,
    xtick = {1, 250, 500, 750, 1000},
    ylabel={Incidence rates},
    ymin = -1.5,
    ymax = 1.5,
    grid style=densely dashed,
    grid=both,
    legend style={
        at={(1, 1)},
        anchor=north east},
    legend columns=1,
    width=127,
    height=130,
    ]
    
    \addplot [blue!80!white, name path = inc-bot, opacity=\contourop, forget plot] table [x=xaxis, y=incid] \prctbot;
    \addplot [blue!70!white, name path = inc-top, opacity=\contourop, forget plot] table [x=xaxis, y=incid] \prcttop;
    \addplot[blue!70!white, fill opacity=\opacity, forget plot] fill between[of=inc-bot and inc-top];
    \addplot[blue, solid] table [x=xaxis, y=incid] {\errtable};

    \draw[dashed, black, line width=1.5pt] (axis cs:350,-1.5) -- (axis cs:350,1.5);
    \draw[dashed, black, line width=1.5pt] (axis cs:700,-1.5) -- (axis cs:700,1.5);

\end{axis}
\end{tikzpicture}
	\end{subfigure}
	\begin{subfigure}{0.24\textwidth}
		\centering
		\begin{tikzpicture}[baseline,scale=1]

\pgfplotstableread{data/realdata/covid_err_med.csv}\errtable
\pgfplotstableread{data/realdata/covid_regret_med.csv}\regrettable

\pgfmathsetmacro{\opacity}{0.3}
\pgfmathsetmacro{\contourop}{0.25}

\begin{axis}[
    xlabel={(b) Time index $t$},
    xmin=1,
    xmax=1060,
    xtick = {1, 250, 500, 750, 1000},
    ymin = 0,
    ymax = .35,
    grid style=densely dashed,
    grid=both,
    legend style={
        at={(0, 1)},
        anchor=north west},
    legend columns=1,
    width=127,
    height=130,
    ]
    
    \addplot[blue, solid] table [x=xaxis, y=err] {\errtable};
    \addplot[red, solid] table [x=xaxis, y=regret] {\regrettable};

    \draw[dashed, black, line width=1.5pt] (axis cs:350,0) -- (axis cs:350,.35);
    \draw[dashed, black, line width=1.5pt] (axis cs:700,0) -- (axis cs:700,.35);

    \legend{Err, Regret}
\end{axis}
\end{tikzpicture}
	\end{subfigure}
    \begin{subfigure}{0.24\textwidth}
		\centering
		 \begin{tikzpicture}[baseline,scale=1]

\pgfplotstableread{data/realdata/finantial_stocks_med.csv}\errtable
\pgfplotstableread{data/realdata/finantial_stocks_prct75.csv}\prcttop
\pgfplotstableread{data/realdata/finantial_stocks_prct25.csv}\prctbot

\pgfmathsetmacro{\opacity}{0.3}
\pgfmathsetmacro{\contourop}{0.25}

\begin{axis}[
    xlabel={(c) Time index $t$},
    xmin=1,
    xmax=1204,
    xtick = {1, 250, 500, 750, 1000},
    ylabel={Stock values},
    ymin = -1.5,
    ymax = 1.5,
    grid style=densely dashed,
    grid=both,
    legend style={
        at={(1, 1)},
        anchor=north east},
    legend columns=1,
    width=127,
    height=130,
    ]
    
    \addplot [blue!80!white, name path = stocks-bot, opacity=\contourop, forget plot] table [x=xaxis, y=stocks] \prctbot;
    \addplot [blue!70!white, name path = stocks-top, opacity=\contourop, forget plot] table [x=xaxis, y=stocks] \prcttop;
    \addplot[blue!70!white, fill opacity=\opacity, forget plot] fill between[of=stocks-bot and stocks-top];
    \addplot[blue, solid] table [x=xaxis, y=stocks] {\errtable};

    \draw[dashed, black, line width=1.5pt] (axis cs:500,-1.5) -- (axis cs:500,1.5);
    \draw[dashed, black, line width=1.5pt] (axis cs:820,-1.5) -- (axis cs:820,1.5);

\end{axis}
\end{tikzpicture}
	\end{subfigure}
    \begin{subfigure}{0.24\textwidth}
		\centering
		 \begin{tikzpicture}[baseline,scale=1]

\pgfplotstableread{data/realdata/finantial_err_med.csv}\errtable
\pgfplotstableread{data/realdata/finantial_regret_med.csv}\regrettable

\pgfmathsetmacro{\opacity}{0.3}
\pgfmathsetmacro{\contourop}{0.25}

\begin{axis}[
    xlabel={(d) Time index $t$},
    xmin=1,
    xmax=1204,
    xtick = {1, 250, 500, 750, 1000},
    ymin = 0,
    ymax = 1.5,
    grid style=densely dashed,
    grid=both,
    legend style={
        at={(0, 1)},
        anchor=north west},
    legend columns=1,
    width=127,
    height=130,
    ]
    
    \addplot[blue, solid] table [x=xaxis, y=err] {\errtable};
    \addplot[red, solid] table [x=xaxis, y=regret] {\regrettable};

    \draw[dashed, black, line width=1.5pt] (axis cs:500,0) -- (axis cs:500,1.5);
    \draw[dashed, black, line width=1.5pt] (axis cs:820,0) -- (axis cs:820,1.5);

    \legend{Err, Regret}
\end{axis}
\end{tikzpicture}
	\end{subfigure}
    \vspace{-1mm}
	\caption{
    Numerical evaluation using real-world data. a) and c) respectively display the median of the standardized incidence rates of COVID-19 and the closing value prices of stocks from S\&P 500. b) and d) plot the error and cumulative regret of the estimated graph when using data from the COVID-19 or the Financial dataset..}\label{fig:experiments2}
    \vspace{-3mm}
\end{figure*}

\subsection{Experiments using real-world data}
Finally, we demonstrate the potential applicability of our method through numerical experiments on dynamic graphs using COVID pandemic and stock market data.
In the absence of a ground truth graph, we compare the online estimate with the offline solution $\bbS_t^*$.

\vspace{2mm}
\noindent
\textbf{Covid-19 experiment.}
The COVID-19 dataset\footnote{\url{https://github.com/CSSEGISandData/COVID-19}} contains the incidence rates related to the COVID-19 outbreak, reported by various U.S. states and territories.
The dataset includes daily reports from April 12th, 2020 to March 9th, 2023.
Nodes represent territories, graph signals encode the reported incidence rates, and the inferred topology connects territories with similar infection patterns.
We start with measurement from 46 nodes, and reports from additional states become available at $t \!=\! 312$ and $t \!=\! 652$.
\cref{fig:experiments2}a shows the standardized time series of incidence rates used to estimate the graph topology, and \cref{fig:experiments2}b presents the error and the average regret of the estimated graph.
We observe a rapid increase in error after t = 350, coinciding with a significant rise in incidence rates (marked by dashed vertical lines) and the arrival of new nodes.
As the slope of the incidence rates decreases, 
Algorithm~1 adapts to the new topology, and the error decreases, demonstrating its ability to recover from disruptions in real-world applications.

\vspace{2mm}
\noindent
\textbf{Financial data experiment.}
We use a dataset comprising the closing price values of 15 stocks (nodes) from different sectors of the S\&P 500.
The data spans from June 1st, 2019, to October 14th, 2022, encompassing the COVID-19 outbreak.
The pandemic had a notorious impact on the economy, as illustrated in \cref{fig:experiments2}c, which plots the median standardized values of the stocks, highlighting the decline and the recovery of the markets.
In this experiment, we start with 13 nodes and introduce 2 new nodes at $t = 800$.
This emulates a situation where new companies appear propelled by the recovery of the economy.
\cref{fig:experiments2}d shows how the error spikes with the sudden market drop due to COVID-19, and again when the market recovers and new nodes are added.
Nevertheless, Algorithm~1 adapts the estimated graph reducing the error, and thus, the average cumulative regret begins to decrease.

\section{Conclusion}\label{s:conclusions}
We addressed the relevant problem of learning the topology of expanding graphs online from streaming data.
The arrival of new nodes poses additional modeling challenges that we overcame by exploiting the block structure of $\bbS_t$ and zero padding.
After designing a covariance update tailored for the arrival of new nodes, we introduced a general online algorithm based on projected proximal gradient descent.
We specialized it to deal with GMRF observations and provided a bound for the dynamic cumulative regret that reflected the impact of incoming nodes.
Finally, we demonstrated its performance using controlled, financial, and COVID-19 data.

\section*{Appendix A: Proof of \cref{t:convergence}}\label{s:thm_proof}

To prove our main result, we rely on the following useful lemma, which is proven in Appendix~B.
\begin{lemma}\label{t:main_bound}
    \vspace{-3mm}
    Let $ \hbS_t$ be the sequence of estimated GSOs procured by Algorithm~\ref{a:nti_expanding} with $\nabla f_t(\bbS)$ and $\Pi_{\ccalS_{N_t}}(\bbS)$ as given in \eqref{e:gradient_ggl} and \eqref{e:projection_ggl}.
    Let $\padd{\bbS^*}{N_t}_t$ be the zero-padded version of $\bbS^*_t$, the unique minimizer of the function $f_t$ over the set $\ccalS_{N_t}$.
   Using a step size $\eta \leq \epsilon^2$, it holds that
    \begin{equation}
        \| \hbS_{t+1} - \padd{\bbS^*}{N_{t+1}}_t \|_F \leq \rho \| \hbS_t - \bbS^*_t \|_F,
    \end{equation}
    where $0 \leq \rho := (1-h\eta/\sigma)^{\frac{1}{2}} < 1$, $\sigma^{-1}$ is the strong convexity constant of $f_t$, and $h \in (0, 1]$.
\end{lemma}

With this in mind, the first step to proof \cref{t:convergence} involves bounding the distance $\| \hbS_t - \bbS_t^* \|_F$ in terms of the distance between $\bbS_t^*$ and $\bbS_{t-1}^*$.
By applying the triangle inequality, we obtain
\begin{align}\label{e:main_bound}
    \| \hbS_t - \bbS_t^* &\|_F \leq \| \hbS_t - \padd{\bbS^*}{N_t}_{t-1} \|_F + \| \bbS_t^* - \padd{\bbS^*}{N_t}_{t-1} \|_F \nonumber \\
    & \leq \rho \| \hbS_{t-1} - \bbS^*_{t-1} \|_F + \| \bbS^*_t - \padd{\bbS^*}{N_t}_{t-1} \|_F,
\end{align}
where the second inequality follows from \cref{t:main_bound}.
Also, recall that $\padd{\bbS^*}{N_t}_{t-1}$ denotes the $N_t \! \times \! N_t$ matrix resulting from zero-padding the 
$N_{t-1} \! \times \! N_{t-1}$ matrix $\bbS^*_{t-1}$, as defined in \eqref{e:zero_padding}.

Next, summing from $t \!= \!1$ to $t \!=\! T$ and using \eqref{e:main_bound} we get
\begin{align}
    \sum_{t=1}^T \| \hbS_t - \bbS^*_t \|_F &\leq \| \hbS_1 - \bbS^*_1 \|_F +  \rho\sum_{t=2}^T \| \hbS_{t-1} - \bbS^*_{t-1} \|_F \nonumber \\
    &+ \sum_{t=2}^T \| \bbS^*_t - \padd{\bbS^*}{N_t}_{t-1} \|_F.
\end{align}
Upon adding and subtracting the term $\rho \| \hbS_T - \bbS^*_T \|_F$ and reordering the terms, we get
\begin{align}
    \sum_{t=1}^T \| \hbS_t &- \bbS^*_t \|_F \leq
    \| \hbS_1 - \bbS^*_1 \|_F - \rho\| \hbS_T - \bbS^*_T \|_F \nonumber \\
    &+  \rho\sum_{t=1}^T \| \hbS_t - \bbS^*_t \|_F +\sum_{t=2}^T \| \bbS^*_t - \padd{\bbS^*}{N_t}_{t-1} \|_F.
\end{align}

Finally, the proof concludes by regrouping the terms and using the constants $K_1$ and $K_2$ defined in \eqref{e:thm_constants}, resulting in 
\begin{equation}
    \sum_{t=1}^T \| \hbS_t - \bbS^*_t \|_F \leq K_1 + K_2 \sum_{t=2}^T \| \hbS_{t-1} - \bbS^*_{t-1} \|_F.
\end{equation}

\section*{Appendix B: Proof of \cref{t:main_bound}}\label{s:main_bound}
This appendix proves \cref{t:main_bound}, an auxiliary result instrumental to obtaining \cref{t:convergence}.
Zero-padding will play a relevant role in addressing the increasing size of the graph, so recall that $\padd{\bbZ}{i}$, previously defined in \eqref{e:zero_padding}, represents an $i \times i$ matrix constructed by zero-padding a smaller matrix $\bbZ$.

To bound $\| \hbS_{t+1} - \padd{\bbS^*}{N_{t+1}}_{t} \|_F$, we first observe that a related result is provided in \cite[Prop. 2]{mokhtari2016online}.
Unfortunately, from the definition of our function $f_t: \reals^{N_t \times N_t} \to \reals$ it follows that the domain of $f_t$ increases with $t$, preventing a direct application of existing results.
To circumvent this limitation, we devise an alternative approach based on a function $\uf_t$ whose domain has a fixed dimension.
Crucially, the sequence of solutions obtained by this alternative algorithm is equivalent to that produced by Algorithm~\ref{a:nti_expanding}.

This proof involves the following key steps: i) design the function $\uf_t$ and establish its relation to $f_t$; ii) formulate the alternative algorithm and link it to Algorithm~\ref{a:nti_expanding}; iii) demonstrate that the alternative algorithm satisfies the conditions from \cite[Prop.~2]{mokhtari2016online}.
These steps are detailed next.

\vspace{2mm}
\noindent\textbf{Designing the function.}
First, we define the set representing the domain of the function.
Consider the set of matrices of size $N_t \times N_t$ zero-padded to obtain $N_T \times N_T$ matrices
\begin{equation}
    \uccalZ = \{ \bbZ \in \reals^{N_T \times N_T } | \bbZ = \padd{\bbY}{N_T} \; \mathrm{for} \; \bbY \in \reals^{N_t \times N_t} \}, \nonumber
\end{equation}
and the set of GSOs of size $N_T$ as defined in \eqref{e:offline_ggl}.
Recall that the zero-padding matrices $\padd{\bbY}{N_T}$ are defined in \eqref{e:zero_padding}, and that $N_T$ denotes the maximum number of nodes of the sequence of graphs.
Then, let the set
\begin{equation}
    \uccalS  :=  \uccalZ \cap \ccalS_{N_T}
\end{equation}
be the set of GSOs of size $N_T \times N_T$ given by a zero-padded GSOs of size $N_t \times N_t$. 
The set is non-empty since zero-padded matrices can also belong to the set $\ccalS_{N_T}$.

The projection onto this set will also be relevant.
For any matrix $\bbZ \in \reals^{N_T \times N_T}$, the projection onto $\uccalS$ is given by
\begin{equation}\label{e:projection_ext}
    \Pi_{\uccalS}(\bbZ) =  
    \begin{bmatrix}
        \Pi_{\ccalS_{N_t}} \left( [\bbZ]_{\ccalV_t} \right) & \bbzero \\
        \bbzero & \bbzero
    \end{bmatrix},
\end{equation}
where the top-left block of size $N_t \times N_t$ is projected into the set $\ccalS_{N_t}$ and the remaining elements are set to 0.

Next, consider the function $\uf_t(\ubS): \uccalS \to \reals$ given by
\begin{align}\label{e:f_ext_eq}
    \uf_t(\ubS) &= \tr\left( \padd{\hbC}{N_T}_t \ubS \right) - \log\det \left( \ubS + \ubI_\epsilon \right) + \alpha d_t(\ubS)  \\
    &=  \tr\left( \hbC_t [\ubS]_{\ccalV_t} \right) - \log\det \left( [\ubS]_{\ccalV_t} + \bbI_\epsilon \right) + \alpha d_t(\ubS), \nonumber
\end{align}
where in a slight abuse of notation we define $d_t(\ubS) := d([\ubS]_{\ccalV_{t-1}}, \hbS_{t-1})$ to match the distance $d$ in \textbf{(AS2)}.
The second equality follows from the properties of the determinant and the block structure of $\ubS$, which come from the domain of the function $\uccalS$, with $[\ubS]_{\ccalV_t}$ being the non-zero (top-left) block of $\ubS$, and
\begin{equation}
    \ubI_\epsilon = \begin{bmatrix}
        \epsilon \bbI_{N_t \times N_t} & \bbzero \\
        \bbzero & \bbI_{(N_T - N_t) \times (N_T - N_t)}
    \end{bmatrix}.
\end{equation}
Here, $\epsilon$ is the positive scalar considered in \eqref{e:offline_ggl}, and $\bbI_{N_t \times N_t}$ is the adjacency matrix of size $N_t \times N_t$, hence specifying the dimensions of the diagonal blocks to avoid confusion.

From \eqref{e:f_ext_eq}, it follows that the value of $\uf_t (\ubS)$ depends only on its block $[\ubS]_{\ccalV_t}$, so $f_t(\bbS) = \uf_t (\padd{\bbS}{N_T})$ for every $\bbS \in \ccalS_{N_t}$.
A direct consequence is that the minimizer 
\begin{equation}
    \ubS_t^* = \argmin_{\ubS \in \uccalS} \uf_t(\ubS) + \lambda \| \ubS \|_1,
\end{equation}
is given by zero-padding $\bbS_t^*$, the minimizer of \eqref{e:offline_ggl}, i.e.,
\begin{equation}\label{e:equivalent_min}
    \ubS_t^* = \padd{\bbS^*}{N_T}_t.
\end{equation}

\vspace{2mm}
\noindent\textbf{Designing the algorithm.}
When a new signal $\bbx_t$ is received a time $t$, update $\hbC_t$ as in \eqref{e:c_update} and compute
\begin{align}
    \dot{\ubS}_t &= \Pi_{\uccalS} \left( T_{\eta\lambda} \left( \hat{\ubS}_{t-1} - \eta \nabla \uf_t(\hat{\ubS}_{t-1}) \right) \right) \label{e:extended_step} \\
    \hat{\ubS}_t &= h \dot{\ubS}_t + (1-h) \hat{\ubS}_{t-1}. \label{e:ext_est}
\end{align}
Note that no zero-padding is required since every $\hat{\ubS}_t$ is a matrix of size $N_T \times N_T$. 
Next, we show that the estimates $\hat{\ubS}_t$ obtained with these operation are given by $\padd{\hbS}{N_T}_t$. 
Put in words, the estimates $\hat{\ubS}_t$ are zero-padded versions [cf. \eqref{e:zero_padding}] of the sequence given by Algorithm~\ref{a:nti_expanding}.

We show this by induction.
Assume $\hat{\ubS}_{t-1} = \padd{\hbS_{t-1}}{N_T}$ for some $t$, and consider the initialization $\hat{\ubS}_{0} = \padd{\hbS_{0}}{N_T}$.
Using $\dot{d}_t(\bbS)$ as a shorthand for $\dot{d}_t(\bbS) := \nabla d([\bbS]_{\ccalV_{t-1}}, \hbS_{t-1})$, the gradient of $\uf_t$ is given by
\begin{equation}
    \nabla  \uf_t  (\padd{\bbS}{N_T}) =
    \begin{bmatrix}
        \hat{\bbC}_t - (\bbS+\epsilon\bbI)^{-1} + \alpha \dot{d}_t(\bbS)  & \!\! \bbzero \\
        \bbzero &  -\bbI \\
    \end{bmatrix},
\end{equation}
where we observe that the top-left block is given by $\nabla_\bbS f_t(\bbS)$.
Because the soft-thresholding is computed entry-wise and based on the projection $\Pi_{\uccalS}$ given in \eqref{e:projection_ext},
it follows that $\dot{\ubS}_t = \padd{\bchkS}{N_T}_t$.
Then, the result follows from noticing that \eqref{e:ext_est} is also computed entry-wise, so $\hat{\ubS}_t = \padd{\hbS_t}{N_T}$.

\vspace{2mm}
\noindent\textbf{Proving the bound.}
Up to this point, we have that
\begin{equation}
    \| \hbS_{t+1} - \padd{\bbS^*}{N_t}_t \|_F = \| \padd{\hbS}{N_T}_{t+1} - \padd{\bbS^*}{N_T}_t \|_F,
\end{equation}
where $\padd{\hbS}{N_T}_{t+1}$ is the zero-padded version of $\hbS_{t+1}$,whixh can be obtained from the steps in \eqref{e:extended_step}-\eqref{e:ext_est}, and $\padd{\bbS^*}{N_T}_t$ is the zero-padded version of $\padd{\bbS^*}{N_t}_t$ as well as the global minimum of $\uf_t$.
Furthermore, note that the proximal operator can be interpreted as an orthogonal projection~\cite{parikh2014proximal}, so the steps in \eqref{e:extended_step}-\eqref{e:ext_est} match those of the method considered in \cite{mokhtari2016online}.
Then, the final ingredient is to show that $\uf_t$ is strongly convex and its gradient is Lipschitz, hence satisfying the conditions in \cite[Prop. 2]{mokhtari2016online}.

To prove the strong convexity of $\uf_t$ and the Lipschitz continuity of $\nabla \uf_t$, we study the Hessian of $\uf_t$, given by
\begin{equation}\label{e:hessian}
    \nabla^2 \uf_t(\ubS) = \left( \ubS + \ubI_\epsilon \right)^{-1} \otimes \left( \ubS + \ubI_\epsilon \right)^{-1},
\end{equation}
with $\otimes$ denoting the Kronecker product.

To show that $\nabla \uf_t$ is Lipschitz continues, it suffices to show that the maximum eigenvalue of $\nabla^2 \uf_t (\ubS)$ is bounded.
From \eqref{e:hessian}, it follows that
\begin{equation}
    \| \nabla^2 \uf_t (\ubS) \|_2 \leq \lambda_{\textnormal{min}}\left( \ubS + \ubI_\epsilon \right)^{-2} \leq \epsilon^{-2},
\end{equation}
where the Lipschitz constant $\epsilon^{-2}$ is independent of $t$.

Similarly, $\uf_t$ is strongly convex if the minimum eigenvalue of its Hessian is strictly larger than 0.
Here, we have
\begin{equation}
    \lambda_{\textnormal{min}} \!\left( \!\nabla^2 \uf_t (\ubS) \!\right) \! \geq \! \lambda_{\textnormal{max}} \left( \ubS \!+\! \ubI_\epsilon \right)^{-2} \!\approx\! \lambda_{\textnormal{max}} \left( \ubS \right)^{-2} = \sigma^{-1} \!\!.
\end{equation}

Since the conditions from \cite[Prop. 2]{mokhtari2016online} are satisfied, with $0 \leq \rho = (1-h\eta/\sigma)^{1/2} < 1$, the result follows from
\begin{align}
    \| \hbS_{t+1} &- \padd{\bbS^*}{N_t}_t \|_F = \| \padd{\hbS}{N_T}_{t+1} - \padd{\bbS^*}{N_T}_t \|_F \nonumber \\
    &\leq \rho \| \padd{\hbS}{N_T}_t - \padd{\bbS^*}{N_T}_t \|_F \nonumber = \rho \| \hbS_t - \bbS^*_t \|_F.
\end{align}





\newpage

\bibliographystyle{IEEEbib}
\bibliography{myIEEEabrv,biblio}

\vfill\pagebreak

\end{document}